\documentclass[10pt,twocolumn,letterpaper]{article}

\usepackage{iccv}              

\usepackage{bm}
\usepackage{array,booktabs}
\usepackage[toc,page]{appendix}
\usepackage{siunitx}  
\usepackage{multirow}
\usepackage[accsupp]{axessibility}  

\newcommand{\x}{\bm{{x}}}
\newcommand{\y}{\bm{{y}}}
\newcommand{\q}{\bm{{q}}}

\newcommand{\n}{\bm{n}}
\newcommand{\co}{\bm{c}}
\newcommand{\norm}[1]{\left\lVert#1\right\rVert}
\newcommand{\I}{\bm{\mathrm{I}}}
\newcommand{\T}{\mathcal{T}}

\newcommand\blfootnote[1]{
    \begingroup
    \renewcommand\thefootnote{}\footnote{#1}
    \addtocounter{footnote}{-1}
    \endgroup
}

\usepackage[acronym]{glossaries}
\newacronym{AP}{AP}{Average Precision}
\newacronym{adaLN}{adaLN}{adaptive layer norm}
\newacronym{BCE}{BCE}{Binary Cross Entropy}
\newacronym{BOP}{BOP}{Benchmark for 6D Object Pose Estimation}
\newacronym{CE}{CE}{Cross Entropy}
\newacronym{CNN}{CNN}{convolutional neural network}

\newacronym{DiT}{DiT}{Diffusion Transformer}
\newacronym{EMA}{EMA}{Exponential Moving Average}
\newacronym{ESA}{ESA}{European Space Agency}

\newacronym{fANOVA}{fANOVA}{functional analysis of variance}
\newacronym{GSO}{GSO}{Google Scanned Objects}
\newacronym{GMM}{GMM}{Gaussian Mixture Model}
\newacronym[plural=GPUs,firstplural=graphics processing units (GPUs)]{GPU}{GPU}{graphics processing unit} 
\newacronym{GAN}{GAN}{generative adverserial network}

\newacronym{HIL}{HIL}{hardware-in-the-loop}
\newacronym{IoU}{IoU}{Intersection over Union}
\newacronym{mAP}{mAP}{Mean Average Precision}
\newacronym{MLP}{MLP}{multi-layer perceptron}
\newacronym{OC-DiT}{OC-DiT}{Object-Conditioned Diffusion Transformer}
\newacronym{ODE}{ODE}{Ordinary Differential Equation}
\newacronym{PnP}{PnP}{Perspective-n-Point}
\newacronym{RoI}{RoI}{Region of Interest}
\newacronym{RANSAC}{RANSAC}{Random Sample Consensus}
\newacronym{SPEC}{SPEC}{Satellite Pose Estimation Challenge}
\newacronym{TPE}{TPE}{Tree-structured Parzen Estimator}
\newacronym{UOIS}{UOIS}{Unseen Object Instance Segmentation}
\newacronym{ViT}{ViT}{Vision Transformer}
\newacronym{VOC}{VOC}{PASCAL Visual Object Classes}
\newacronym{VAE}{VAE}{variational autoencoder}
\newacronym{yolo}{YOLO}{You Only Look Once}
\newacronym{ZSI}{ZSI}{Zero-Shot Instance Segmentation}

\definecolor{iccvblue}{rgb}{0.21,0.49,0.74}
\usepackage[pagebackref,breaklinks,colorlinks,allcolors=iccvblue]{hyperref}

\def\paperID{9815} 
\def\confName{ICCV}
\def\confYear{2025}

\title{Conditional Latent Diffusion Models for Zero-Shot Instance Segmentation}

\author{
{Maximilian Ulmer$^{1, 2}$ \hspace{0.25em} Wout Boerdijk$^{1, 3}$ \hspace{0.25em} Rudolph Triebel$^{1, 2}$ \hspace{0.25em} Maximilian Durner$^{1, 3}$}
\vspace{0.2em}
\and {$^{1}$German Aerospace Center (DLR)}
\hspace{-2em}
\and {$^{2}$Karlsruhe Institute of Technology}
\and {$^{3}$Technical University of Munich}
}

\begin{document}
\maketitle
\begin{abstract}
This paper presents \acrfull{OC-DiT}, a novel class of diffusion models designed for object-centric prediction, and applies it to zero-shot instance segmentation. 
We propose a conditional latent diffusion framework that generates instance masks by conditioning the generative process on object templates and image features within the diffusion model's latent space. 
This allows our model to effectively disentangle object instances through the diffusion process, which is guided by visual object descriptors and localized image cues. 
Specifically, we introduce two model variants: a coarse model for generating initial object instance proposals, and a refinement model that refines all proposals in parallel. 
We train these models on a newly created, large-scale synthetic dataset comprising thousands of high-quality object meshes. 
Remarkably, our model achieves state-of-the-art performance on multiple challenging real-world benchmarks, without requiring any retraining on target data. 
Through comprehensive ablation studies, we demonstrate the potential of diffusion models for instance segmentation tasks. 
Code is available at \href{https://github.com/DLR-RM/oc-dit}{https://github.com/DLR-RM/oc-dit}.
\end{abstract}
\blfootnote{Corresponding author: \href{mailto:maximilian.ulmer@dlr.de}{maximilian.ulmer@dlr.de}}
\vspace{-1em}    
\section{Introduction}
\label{sec:intro}
\begin{figure}
  \centering
   \includegraphics[width=\linewidth]{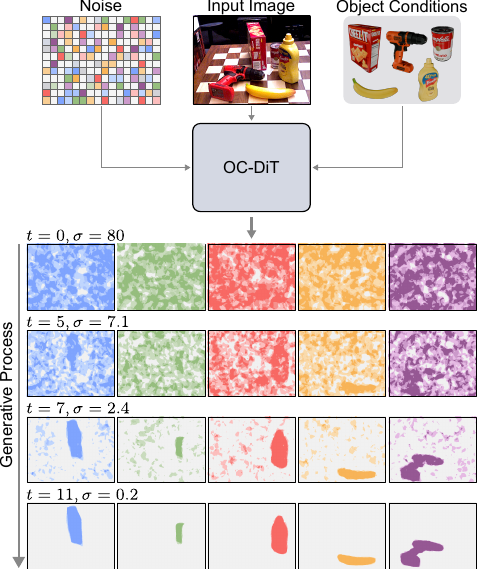}
   \caption{\textbf{Conditional Generative Process for Zero-Shot Instance Segmentation.} The \acrfull{OC-DiT} is a generative model that conditions the sampling process on visual descriptors of query objects such that only these objects are considered. 
   It is trained on thousands of object meshes and millions of samples and can be rapidly deployed to new object instances without retraining.}
   \label{fig:onecol}
   \vspace{-0.8em}
\end{figure}

Instance segmentation is a core computer vision task that detects and separates object instances at the pixel level.
As an initial step in many vision-based applications, it supports 6D pose estimation~\cite{lin_hipose_2024,ulmer20236d,hodan_bop_2024}, tracking~\cite{stoiber_iterative_2022}, shape completion~\cite{yan_shapeformer_2022,humt_shape_2023}, or robotic applications (\eg grasping~\cite{sundermeyer_contact_2021,newbury2023deep}).
In literature, two distinct directions at opposite ends of the spectrum exist.
One focuses on well-established 2D object instance segmentation approaches~\cite{he_mask_2017,carion_end_2020}, where models are trained on annotated real or synthetic datasets with predefined target objects.
By learning instance-specific features, they achieve reliable performance but require all target objects during training. 
Integrating new instances demands retraining, making adaptation prohibitively expensive and time-consuming.
At the other end, there are class-agnostic methods~\cite{xie2021unseen,durner2021unknown,xiang2021learning}. 
These approaches learn a generic definition of objects, allowing them to segment a varying group of arbitrary instances without prior knowledge. 
They operate without additional training but lack semantic information and the ability to assign specific labels to their predictions.

This work aims to bridge both concepts by addressing \gls{ZSI}, which produces class-specific instance predictions of arbitrary objects while only demanding conditioning information during inference.
Thus, unlike traditional methods which require the CAD model of the target objects at training time, \gls{ZSI} retains instance-specific predictions without model-specific training.
Generally, \gls{ZSI} methods generate feature representations of rendered template images based on the related object model at test time.
These are then compared to the real image features to retrieve the objects of interest in a query image.
Besides the induced sim-to-real gap, the limited amount of templates prohibits the analysis of the changing appearance of a target object in a large number of scales.

Recent works tackle \gls{ZSI} in two steps: extracting region proposals and comparing them to template features.
In contrast, we introduce \textit{\gls{OC-DiT}}, which integrates object hypothesis generation and feature matching in an end-to-end manner.
Our diffusion-based approach separates individual instances through a generative process (see \cref{fig:onecol}), matching template and image features in its latent space.
This introduces diffusion as a promising novel concept for instance segmentation-related tasks.

To summarize, our contributions are:
\begin{enumerate*}[label=(\roman*)]
    \item \acrshort{OC-DiT}, a novel class of diffusion models designed for object-centric predictions, built on the transformer architecture to achieve high flexibility, adapting to different input resolutions, template counts, and target objects.
    \item Leveraging this flexibility, we introduce two variants of our model for initial estimates in high-resolution images and subsequent region of interest refinement.
    \item Our method, trained exclusively on synthetic data, performs competitively against state-of-the-art methods for model-based 2D segmentation of unseen objects.
    \item We contribute two large synthetic datasets to advance research on related topics.
\end{enumerate*}

\section{Related Work} 
\label{sec:related_work}
\noindent\textbf{Zero-Shot Instance Segmentation}.
To generalize from seen to unseen object classes, seminal works build up on word-vectors to establish a visual-semantic mapping~\cite{zheng2021zero} or exploit unlabeled videos~\cite{du2021learning}.
\acrfull{UOIS} is often confined to table-top scenarios, and respective works employ RGB-D modalities in a two-staged approach~\cite{xie2021unseen} or specific clustering losses~\cite{xiang2021learning}.
Durner~\etal~\cite{durner2021unknown} make use of stereo imagery to cope with reflective or texture-less object material properties.

The perception of unseen objects has also been addressed by designated tasks in the \gls{BOP}~\cite{hodan_bop_2024} challenge, where the desired target object is described by its CAD model.
The instance segmentation sub-task is then commonly solved by matching rendered images of the target with a query image.
Here, recent works often build upon advances in promptable Foundation Models~\cite{lin2024sam,zhao2023fast,oquab2023dinov2,liu2024grounding}.
For instance, CNOS~\cite{nguyen2023cnos} describes rendered templates from CAD models by DINOv2~\cite{oquab2023dinov2}~\emph{cls} tokens. 
SAM~/~FastSAM~\cite{lin2024sam,zhao2023fast} generates semantic region proposals, and object instance masks are generated by matching every proposal against the template tokens.
The problem of SAM generating region proposals rather than object proposals is addressed in NIDS-Net~\cite{lu2024adapting}, who prompt Grounding DINO with \emph{objects} and subsequently segments the detections with SAM. 
While addressing the task of CAD-based 6D Pose Estimation, both SAM6D~\cite{lin2024sam} and ZeroPose~\cite{chen2023zeropose} generate object instance segmentation masks as intermediate solutions: 
The former calculates object matching scores between SAM proposals and rendered templates based on semantics, appearance, and geometry, while for the matching to a query image the latter extracts specific visual embeddings from CAD renderings with DINOv2 and dedicated geometrical features from the raw CAD model.

\vspace{0.4em}\noindent\textbf{Diffusion Models for Segmentation}. Diffusion models have originally been proposed for generative tasks, with seminal works of Ho~\etal~\cite{ho2020denoising}, Dhariwal~\etal~\cite{dhariwal2021diffusion}, and Rombach~\etal~\cite{rombach2022high} showcasing their competitiveness in terms of image synthesis to Generative Adversarial Networks.
In the context of segmentation, a significant advantage given the inherent stochastic sampling process is the possibility to generate multiple segmentation proposals, resulting in pixel-wise uncertainty maps.
For instance,~\cite{amit2021segdiff} sums up the features of an encoded query image and the current noise, and decodes it into a segmentation mask. 
\cite{tan2023diffss} employ diffusion models for few-shot segmentation, and~\cite{wolleb2022diffusion, rahman2023ambiguous} explore diffusion for segmentation of medical images.

\section{Method}
\label{sec:method}
\begin{figure*}
  \centering
\includegraphics[width=\linewidth]{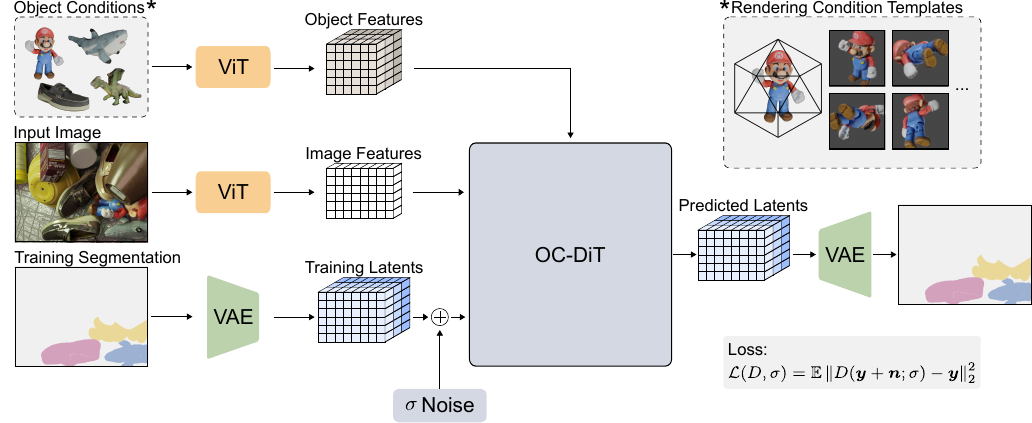}
\caption{\textbf{Architecture Overview.} Our model consists of three learned components: a pre-trained \acrshort{ViT} for feature extraction, a pre-trained \acrshort{VAE} that compresses binary segmentation masks into a lower-dimensional latent space, and the \acrshort{OC-DiT} transformer. The objects condition the predictions of the diffuser, effectively guiding the generative process. We render templates from an icosphere around each object and extract features from these templates. The \acrshort{VAE} decoder is only used during inference.}
\vspace{-1.2em}
\label{fig:oc-dit_architecture}
\end{figure*}
This section details the \acrshort{OC-DiT} architecture (\cref{fig:oc-dit_architecture}) and its instance segmentation diffusion process. We first review the diffusion formulation (\cref{subsec:prelims}) before presenting the core contribution: the \acrshort{OC-DiT} transformer latent diffusion model (\cref{subsec:oc_dit}, \cref{fig:oc-dit_transformer}).



\subsection{Preliminaries}
\label{subsec:prelims}
\noindent\textbf{Diffusion Models}~\cite{Sohl-Dickstein2015-qp} are probabilistic models designed to generate samples from a distribution $p_{\text{data}}(\x)$, with standard deviation $\sigma_{\text{data}}$. 
The core idea is to train a denoising function $D(\x;\sigma)$ that minimizes the expected $L_2$ error for samples drawn from $p_{\text{data}}(\x)$ for noise $\sigma \in [0, \sigma_{\text{max}}]$, \ie,
\begin{equation}\label{eq:diffusion_target}
    \mathcal{L}(D;\sigma) = \mathbb{E}_{\y, \n } \norm{D(\y + \n; \sigma) - \y}_{2}^{2}
\end{equation}
for clean training samples $\y\sim p_{\text{data}}$ and Gaussian noise $\n \sim \mathcal{N}(0,\sigma^{2}\I)$~\cite{karras2022elucidating}. 
At inference, we sample pure noise $\x_N \sim \mathcal{N}(0,\sigma^{2}_{\text{max}}\I)$, with $\sigma_{\text{max}} \gg \sigma_{\text{data}}$, and sequentially apply $D$ to denoise it with noise levels $\sigma_N = \sigma_{\text{max}} >\dots >  \sigma_0 = 0$ that each noise level $\x_i \sim p(\x_i;\sigma_i)$.
After $N$ steps, a noise-free data sample $\x_0 \sim p(\x_0;0)$ is revealed. 

A probability flow \acrfull{ODE}~\cite{song2020score} is required to evolve a sample towards lower noise levels. 
Following~\cite{karras2022elucidating, song2020score} and $\sigma(t)=t$, the evolution $\x \sim p(\x;\sigma)$ with respect to a change in $\sigma$ is described by 
\begin{equation}
    \mathop{d\x}=-\sigma \nabla_{\x} \log p(\x;\sigma) \mathop{d\sigma}
\end{equation}
with the score function $\nabla_{\x} \log p(\x;\sigma)$~\cite{hyvarinen2005estimation}. 
Given $D(\x;\sigma)$ that minimizes (\ref{eq:diffusion_target}), the score function is 
\begin{equation}
    \nabla_{\x} \log p(\x;\sigma) = \big( D(\x;\sigma) - \x \big) / \sigma^2 .
\end{equation}
We can approximate this vector field using a neural network $D_\theta(\x;\sigma)$ parameterized by weights $\theta$, such that
\begin{equation}
    \mathop{d\x} / \mathop{d\sigma} = - \big( D_\theta(\x;\sigma) - \x \big) / \sigma
\end{equation} and we can use $D_\theta(\x;\sigma)$ to step through the diffusion process. For more information, please refer to \cite{karras2022elucidating}.

\vspace{0.3em}\noindent\textbf{Preconditioning, Loss Weighting, and Discretization}. 
Previous work~\cite{song2020denoising, nichol2021improved} has shown that directly training $D_\theta(\x;\sigma)$ is difficult, as the magnitude of the input signal $\x=\y+\n$ varies immensely dependent on the noise $\n \sim \mathcal{N}(0,\sigma^{2}\I)$. 
We follow \cite{karras2022elucidating} and do not train $D_\theta$ directly, but a network $F_\theta$:
\begin{equation} \label{eq:denoiser_network}
    D_\theta(\x,\sigma) = c_\text{skip}(\sigma)\x + c_\text{out}( \sigma)F_\theta \big(c_\text{in}(\sigma)\x;c_\text{noise}(\sigma) \big).
\end{equation}
In this setting, there is a $\sigma$-dependent skip connection $c_\text{skip}(\sigma)$, such that we train the model to estimate a mixture of signal $\x$ and noise $\n$. $c_\text{in}(\sigma)$ and $c_\text{out}(\sigma)$ scale the input and output magnitudes, and $c_\text{noise}(\sigma)$ maps the noise level to a conditioning value.

Weighing the loss by $\lambda(\sigma)=1/c_\text{out}(\sigma)^2$, we can equalize the contribution of different noise levels at initialization~\cite{karras2024analyzing}, such that $\lambda(\sigma)\mathcal{L}(D_\theta;\sigma)=1$.
Interestingly, this setting is equivalent to multi-task learning where each task corresponds to optimizing $\mathcal{L}_i(D_\theta, \sigma_i)$.
In its simplest formulation, these losses can be weighted uniformly~\cite{rombach2022high}.
In contrast, \cite{karras2024analyzing, kendall2018multi} propose weighting each contribution by the uncertainty $u(\sigma_i)=\ln{\sigma_i^2}$ of the model about the task $i$, leading to the loss:
\begin{equation}
    \mathcal{L}(D_\theta,u)=\mathbb{E}_\sigma \bigg[\frac{\lambda(\sigma)}{e^{u(\sigma)}} \mathcal{L}(D_\theta;\sigma) + u(\sigma) \bigg].
\end{equation}
In practice, $u(\sigma)$ is implemented through a small neural network trained together with $D_\theta$.

Finally, during inference we want to step backwards from $\x_N$ to $\x_0$ in $N$ time steps $\{t_{i\in{0,\dots,N-1}}\}$. 
We condition the model on the current continuous noise level $\sigma$ instead of discrete time steps; hence, we need to discretize $\{ \sigma_i \}$.
We follow the discretization of \cite{karras2022elucidating}, such that the $i$th sampling step is given by
\begin{equation}
    \sigma_{i} = \bigg( \sigma_\text{max}^{\frac{1}{\rho}} + \frac{i}{N-1} \big( \sigma_\text{min}^{\frac{1}{\rho}} -  \sigma_\text{max}^{\frac{1}{\rho}} \big) \bigg)^\rho .
\label{eq:discretization}
\end{equation}
Here, $\rho$ becomes an important parameter that controls the magnitude of $\sigma_i$ in consecutive time steps.

\subsection{Object-Conditioned Diffusion Transformer}
\label{subsec:oc_dit}
\begin{figure*}
  \centering
\includegraphics[width=\linewidth]{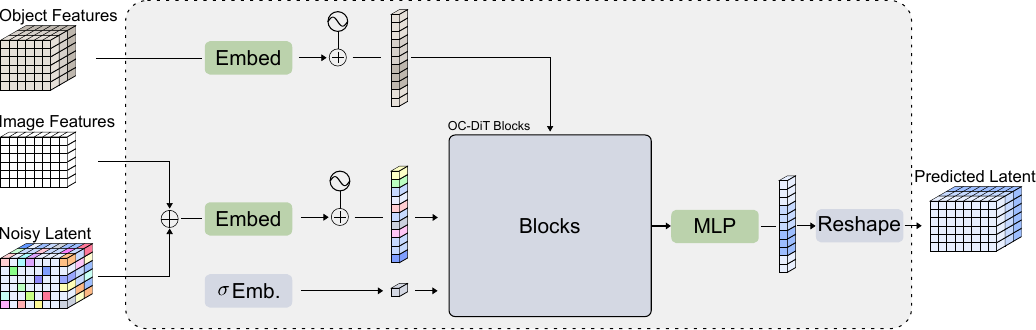}
\caption{\textbf{\acrshort{OC-DiT} Components.} 
The transformer blocks take three inputs: query tokens (concatenated image features and noisy latents with learned positional encoding), embedded template tokens, and embedded noise levels. An \acrshort{MLP} projects the transformer output to the final output dimension.
}
\vspace{-1.2em}
\label{fig:oc-dit_transformer}
\end{figure*}
We introduce the \acrfull{OC-DiT}, inspired by the \acrfull{DiT}~\cite{peebles2023scalable}, designed initially to generate RGB images. 
Although it supports conditioning on time steps and class labels, it was not designed for large-scale conditioning (\eg object features), which our approach enables.

Given a pair $\co=(\bm{I},\mathcal{O})$ of an input image $\bm{I} \in \mathbb{R}^{H \times W \times 3}$, with $H$ and $W$ being the resolution parameters, and a set of target objects $\mathcal{O}=\{o_1,\dots,o_{N_\mathcal{O}}\}$, we pose instance segmentation as a conditional generative process.
Our goal is to sample the conditional distribution $p \big( \x | \co \big)$. 
At inference, we condition the outcome of $\x$ to segment our target objects $\mathcal{O}$ by evolving the distribution $p \big( \x | \co  ; \sigma_{max} \big)$ to $p \big( \x | \co  ; \sigma=0 \big)$. 
We achieve this by training the denoising network $D_\theta(\x;\sigma,\co)$~(\ref{eq:denoiser_network}) that accepts conditioning inputs. 

We design \acrshort{OC-DiT} for flexibility across input resolutions, object count $N_\mathcal{O}$, and templates per object $N_\mathcal{T}$.
Therefore, we adopt \gls{DiT}, based on \acrfull{ViT}, due to its patch-based sequence processing. 
\cref{fig:oc-dit_transformer} depicts an overview of our method.
It comprises embedding layers for input tokenization, the proposed \acrshort{OC-DiT} transformer blocks, and a projection layer for output mapping.

\vspace{0.3em}\noindent\textbf{Latent Diffusion of Bernoulli Distributions}. 
In principle, we could directly sample the distribution of binary segmentation masks. 
However, formulation of Karras~\etal~\cite{karras2022elucidating,karras2024analyzing} considers normally distributed RGB images, with $\mu_\text{data}=0$, and $\sigma_\text{data}=0.5$, 
These requirements differ significantly from Bernoulli-distributed binary segmentations. 
Adding Gaussian noise to binary data in the forward process does not yield normally distributed training data $p_{\text{data}}$.
To address this, we use a $\beta$-\acrfull{VAE} to compress masks into a low-dimensional latent space and perform latent diffusion~\cite{rombach2022high}.
The \acrshort{VAE} allows shaping the latent space statistics during training to align with the requirements. 
Furthermore, the reduced dimensionality drastically improves computational efficiency.

Hence, $\x \in \mathbb{R}^{h \times w \times d}$ represents a latent code, with height $h$, width $w$, and channels $d$. This code is generated by an encoder $V_\text{enc}$ from a binary image $\bm{b} \in \{0,1\}^{H \times W}$, resulting in $\x=V_\text{enc}(\bm{b})$. 
The encoding process effectively downsamples the image by a factor of $H/h$. 
Conversely, a decoder $V_\text{dec}$ reconstructs a confidence map $\bm{\tilde{b}} \in [0,1]^{H \times W}$ for each pixel position, where $\bm{\tilde{b}}=V_\text{dec}(V_\text{enc}(\bm{b}))$. 

Using the \acrshort{VAE}, the diffusion process solely takes place in the latent space. 
During training, $V_\text{enc}$ generates latent codes $\x$ from binary masks $\bm{b}$, which are then corrupted with noise (see \cref{fig:oc-dit_architecture}). 
We train the denoiser $D_\theta$ to recover the original codes, such that $V_\text{dec}$ is not needed during training.
Conversely, during inference, pure noise is iteratively denoised by $D_\theta$ over $N$ steps. 
$V_\text{dec}$ is only applied to the final denoised latent code $\x_0$ and $V_\text{enc}$ is not required.

\vspace{0.3em}\noindent\textbf{Query Embeddings}. 
We generate one query $\q$ for each target object we aim to segment.
Each query consists of a latent code concatenated with the features of a pre-trained backbone $f_{\bm{I}}$.
The origin of the latent code depends on the stage. 
During training it is a ground truth latent $\x = V_\text{enc}(\bm{b})$ corrupted by noise $\y = \x + \n$, with $\n \sim \mathcal{N}(0,\sigma^{2}\I)$.
Yet, during inference, the latent predicted in the previous time step $\x_{i-1}$ is used.

After concatenation, a convolutional layer embeds each query into the embedding space of dimensionality $d_e$, as in the original \acrshort{ViT}~\cite{dosovitskiy2020image}. 
After flattening the queries along the spatial dimensions to $n_{\q} = h \cdot w$, we obtain query tokens $\tau^{\q} =[\tau^{\q}_0, \dots, \tau^{\q}_{N_\mathcal{O}}] \in \mathbb{R}^{N_\mathcal{O} \times n_{\q} \times d_e}$ for all objects.
Instead of \acrshort{DiT}'s frequency-based positional encoding, we use a learned encoding for 2D patches and 1D object positions.

\vspace{0.3em}\noindent\textbf{Conditioning Objects Template Embeddings}.
Our goal is to condition the diffusion process on the set of object instances $\mathcal{O}$.
Similar to previous work~\cite{nguyen2023cnos, lin2024sam}, we render a set of $N_\mathcal{T}$ templates $\mathcal{T}_j$  for each object, resulting in an overall set of templates $\mathcal{T}=\{\mathcal{T}_1,\dots,\mathcal{T}_{N_\mathcal{O}}\}$. 

We extract features for each template using a pre-trained \acrshort{ViT}~\cite{dosovitskiy2020image}, resulting in a lower-dimensional feature space spatially defined by the model's patch size and embedding dimensionality. 
Next, a convolutional layer embeds each template before being flattened to $n_\T = h_\T \times w_\T $ tokens per template, to the shape $\tau_\T \in \mathbb{R}^{N_\mathcal{O} \times N_\T \times n_\T \times d_e}$. Finally, we apply a learned positional encoding to the templates, encoding the 2D patch position, 1D object position, and 1D template position.

\vspace{0.3em}\noindent\textbf{Noise Embeddings}. 
We follow \cite{karras2022elucidating} and do not condition the network on the time step in the diffusion process, but the logarithm of the current noise level $c_{\text{noise}}(\sigma)$. 
We use the magnitude-preserving Fourier features~\cite{karras2024analyzing} to embed the values into the embedding dimensionality before being passed through a linear layer and a SiLU~\cite{hendrycks2016gaussian} layer.

\begin{figure}
  \centering
   \includegraphics[width=0.8\columnwidth]{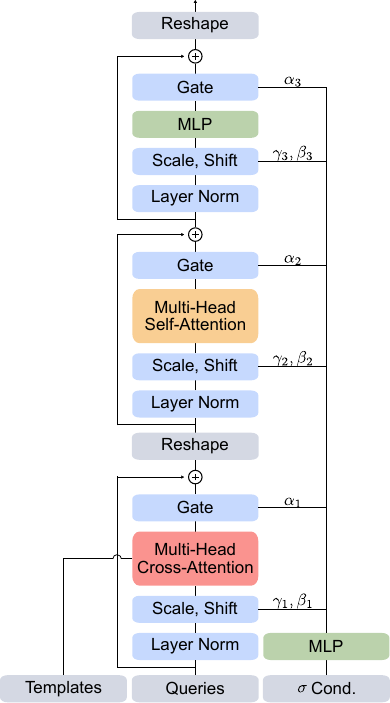}
   \caption{\textbf{OC-DiT Transformer Block.} During the cross-attention phase, each query can only attend to the template tokens of its assigned object. After cross-attention, we reshape the query tokens such that during the self-attention phase, each token can attend to all other query tokens.}
   \label{fig:ocdit_block}
\vspace{-1.2em}
\end{figure}

\vspace{0.3em}\noindent\textbf{OC-DiT Decoder}. 
After the embedding, a sequence of transformer blocks processes the query tokens $\tau^{\q}$.
\cref{fig:ocdit_block} shows the components of each block.
It consists of three layers: a cross-attention layer and a self-attention layer, followed by a \acrfull{MLP}. 
Importantly, we rearrange the query tokens in the cross-attention phase of the block such that each query only attends to template tokens of its assigned object. 
After the cross-attention layer, we reshape the query tokens such that each query token attends to all other query tokens during self-attention.
This reshaping significantly reduces the computational cost of the cross-attention stage and is crucial for the model to learn that each query commits to one object.
Similar to \acrshort{DiT}, we employ an \acrfull{adaLN}~\cite{perez2018film} in the self-attention and \acrshort{MLP} block and add a third \acrshort{adaLN} to the cross-attention part. 
The goal of \acrshort{adaLN} is to modulate tokens by scaling $\gamma$, shifting $\beta$, and gating $\alpha$, depending on the noise level. 
The scale and shift are applied directly after the layer normalization, and the gating is used immediately before the residual connection.
The values $\gamma, \beta$, and $\alpha$ are regressed by passing the noise level through a linear layer.
The weights of the \acrshort{adaLN} layer are initialized to zero, as proposed in \cite{peebles2023scalable}.
After passing through the \acrshort{MLP} stage of the block, the queries are again reshaped to their original shape.
Finally, after the transformer blocks, a final \acrshort{MLP} projects the query tokens from the embedding space to the latent dimensionality. 

\vspace{0.3em}\noindent\textbf{Training and Inference}.
During training, we draw samples from our dataset consisting of pairs of images $\bm{I}$, segmentation masks $\bm{b}$, and corresponding objects $\mathcal{O}$. 
If there are more than $N_\mathcal{O}$ objects in a sample, we randomly select $N_\mathcal{O}$ objects weighted by their proximity to a sampled foreground point.
This selection has the effect that selected objects are biased to be close to each other, resulting in a higher number of pixels that belong to a foreground object per sample.
We hypothesize that training on samples with mainly background pixels can deteriorate learning performance. 
In the forward pass of the model, we randomly sample noise levels $\sigma \in [\sigma_\text{min}, \sigma_\text{max}]$ for each training sample.
We apply the noise to the ground truth latent codes $\x = V_\text{enc}(b)$, such that $\y = \x + \n$, and train the model to predict $\x$. 

At inference, we aim to generate instance segmentations from pure noise. 
We employ the stochastic sampler proposed in \cite{karras2022elucidating}.
At inference, we start the diffusion process at timestep $N$ by sampling noise $\x_N \sim \mathcal{N}(0,\sigma^{2}_{\text{max}}\I)$ that we input to our denoiser $D_\theta$ as queries $\q = \x_N \oplus f_{\bm{I}}$.
The sampler works by raising the noise level at each time step $i$ and then stepping through the \acrshort{ODE} to yield $\x_{i-1}$. 
We repeat this process for $N$ steps until we reach $\x_0$.

\vspace{0.3em}\noindent\textbf{Test-Time Ensembling}. 
The stochastic nature of the diffusion process results in varying predictions based on the initial noise $\x_N \sim \mathcal{N}(0,\sigma^{2}_{\text{max}}\I)$ of the process.
We exploit this property by averaging $\bm{\tilde{b}}$ across multiple inference passes, effectively sampling from $p \big( \x | \co \big)$ multiple times.
Optionally, we can add spatial test-time augmentations, like scaling and translation.
Hence, for an initial ensemble of size $N_{\text{ensemble}}$, we augment each of them $N_\text{aug}$ times, resulting in a total of $N_{\text{ensemble}} \times N_\text{aug}$ forward passes per test sample.
Finally, we align all predictions to aggregate the final ensemble prediction.

\section{Experiment}
\label{sec:experiment}
This section covers implementation details, compares our method to the state-of-the-art, and ablates design choices.
\subsection{Implementation}
\label{subsec:implementation}
\begin{figure*}
  \centering
\includegraphics[width=0.9\linewidth]{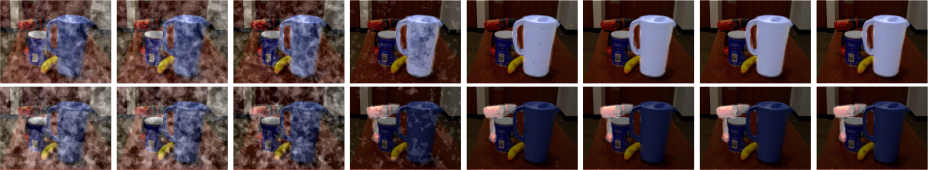}
\caption{\textbf{Visualization of object-conditioned diffusion process} of the \gls{OC-DiT} exemplarily shown for two YCB-V objects in the scene.}
\vspace{-1.2em}
\label{fig:viz_max}
\end{figure*}

\noindent\textbf{Diffuser Setup}.
We use a pre-trained, frozen DINOv2~\cite{oquab2023dinov2} backbone in its small configuration, with a patch size of $14$, to generate features for the input image and templates.
Our \acrshort{VAE} implementation closely follows Stable Diffusion v2~\cite{rombach2022high}.
We standardize the latent space to $\sigma_{\text{data}}=0.5$, as proposed in~\cite{karras2024analyzing}.
During training of \acrshort{OC-DiT}, we use a log-uniform noise distribution with $\sigma \in [0.002, 80]$, unlike the normal distributions used in \cite{karras2022elucidating, karras2024analyzing}.
For all experiments, except specified differently, we use $18$ diffusion steps during inference with a discretization defined by $\rho=15$.

\vspace{0.3em}\noindent\textbf{Model Variants Training Setup}.
Our model can process varying object counts $N_\mathcal{O}$, template numbers $N_\mathcal{T}$, and input resolutions. 
This flexibility allows us to train two different models: \texttt{coarse} and \texttt{refine}.
\texttt{coarse} first predicts segment proposals for the entire input image, conditioned on multiple objects. 
We use these proposals to calculate modal bounding boxes for the subsequent refinement stage. 
Unlike the refinement model, the coarse model is trained exclusively with samples containing many true positives, biasing it towards positive estimations.
\texttt{refine} then only generates segmentations on the lower-resolution modal proposal region, conditioned on just one object.
During training, we expose the refinement model to samples that may include false positives (\ie, the object of interest is not in the crop).
This setup trains the model to discriminate between true and false positives, optimizing for high precision.
While a single model could theoretically perform both stages, we found that dedicated weights for each stage provide superior performance.
Although the refinement stage significantly increases computation, it substantially improves segmentation accuracy, especially for challenging scenes, as shown in \cref{subsec:bop_results}.

\vspace{0.3em}\noindent\textbf{Training Data Generation}.
A diverse training dataset is required to succeed in the zero-shot domain and extrapolate to new scenes and objects. To this end, we train our models on three different datasets: 
\begin{enumerate*}[label=(\roman*)]
    \item a dataset for 6D pose estimation~\cite{labbe2022megapose} using \acrfull{GSO}~\cite{downs2022google} meshes containing around $1$ million samples, \label{item:megapose}
    \item a new dataset using \acrshort{GSO} meshes aimed at segmentation, containing more objects per training sample, and \label{item:obd_gso}
    \item a new dataset using $2600$ Objaverse~\cite{deitke2023objaverse} meshes aimed at segmentation from a large variety of object classes and origins. \label{item:obd_objaverse}
\end{enumerate*}
All three datasets are purely synthetic, rendered with BlenderProc~\cite{denninger2019blenderproc}.
We randomly select $12$ to $30$ instances per scene and render around $500$ thousand samples, totaling approximately 1 million new data samples.

\vspace{0.3em}\noindent\textbf{Evaluation Metric}.
We evaluate our method using the \acrfull{AP} metric used for the COCO and \gls{BOP} challenge evaluations~\cite{sundermeyer2023bop}. The AP metric is the mean of precision values at \acrfull{IoU} thresholds ranging from $50\%$ to $95\%$ with a step size of $5\%$. 

\subsection{BOP Results}
\label{subsec:bop_results}
\begin{table}
\vspace{-1.2em}
\centering
\begin{tabular}{|lcccc|}
\hline
\acrshort{AP} & YCBV & TUDL & LMO & HB\\
\hline
CNOS~\cite{nguyen2023cnos} & 59.9 & 48.0 & 39.7 & 51.1\\
SAM6D~\cite{lin2024sam} & 60.5 & 56.9 & 46.0 & 59.3\\
NIDS~\cite{lu2024adapting} & 65.0 & 55.6 & 43.9 & 62.0\\
MUSE & 67.2 & 56.5 & \textbf{47.8} & 59.7\\
LDSeg & 64.7 & 58.7 & \textbf{47.8} & \textbf{62.2}\\
\hline
Ours \texttt{coarse} & 68.6 & 32.5 & 29.6 & 52.4 \\
Ours \texttt{refined} & \textbf{71.7} & \textbf{59.4} & 40.1 & 61.5 \\
\hline
\end{tabular}
\caption{\textbf{Results on the \acrshort{BOP} challenge datasets.} We report the \acrshort{AP} for the task of model-based 2D segmentation of unseen objects. The highest value in each column is denoted in \textbf{bold}.}
\vspace{-1.5 em}
\label{tab:bop_results}
\end{table}
We demonstrate the capabilities of OC-DiT on the highly adopted BOP benchmark\footnote{Values from \href{https://bop.felk.cvut.cz/leaderboards/segmentation-unseen-bop23/bop-classic-core/}{BOP Challenge}, visited on 27.02.2025} for the task of model-based 2D segmentation of unseen objects.
At test time, the method receives an RGB image that shows objects unseen during training in an arbitrary number of instances, all from one specified dataset.
We present the performance of our method's results alongside state-of-the-art in \cref{tab:bop_results}.
We use the same model, without additional fine-tuning on the target data, for all benchmarks trained on $N_\mathcal{O}=8$ and evaluated with $N_\mathcal{O}=12$. 
As conditioning objects, we consider each dataset's ground truth objects and fill the remaining positions with random distractor objects from the same dataset.

For the \texttt{coarse} estimation, we apply $N_{\text{ensemble}} = 8$ and $N_\text{aug} = 5$.
Finally, we add a sixth spatial test-time augmentation guided by the estimated spatial confidence values of the previous results.
For the refinement stage, we select results with a confidence higher than $0.2$, calculated by averaging predicted foreground pixels.
For the YCBV~\cite{xiang2017posecnn} and TUDL~\cite{hodan_bop_2024} datasets, we keep the input image resolution to the training resolution. 
In contrast, we found that increasing the input resolution to $420 \times 560$ and interpolating patch positional encodings improves performance for HB~\cite{kaskman2019homebreweddb} and LMO~\cite{brachmann2014learning}.
Next, the \texttt{refine} model uses $N_{\text{ensemble}} = 8$ and $N_\text{aug} = 5$, consisting of translated and scaled crops.
We select results with a confidence higher than $0.5$ for the final submission.
As shown in \cref{tab:bop_results}, our method achieves strong results on YCBV, TUDL, and HB. 
On YCBV, already, the \texttt{coarse} model reaches state-of-the-art.
The performance on the LMO data we attribute to biases in our training data towards larger objects.
We also observe a positive impact of the \texttt{refine} model, especially for TUDL and LMO.
A possible cause is a high quantity of false positives estimated by the \texttt{coarse} model, which \texttt{refine} can correct.

\begin{figure*}
  \centering
    \begin{subfigure}{0.33\linewidth}
    \includegraphics[width=\linewidth]{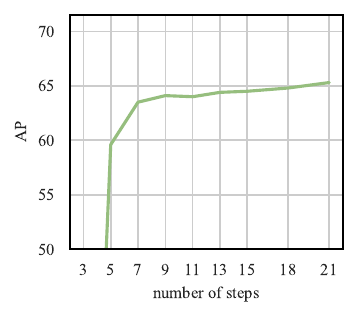}
    \caption{Diffusion Inference Steps}
    \label{fig:diffusion_steps}
  \end{subfigure}
  \hfill
  \begin{subfigure}{0.33\linewidth}
    \includegraphics[width=\linewidth]{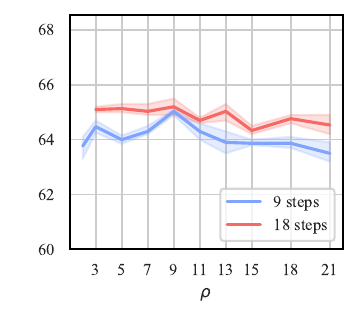}
    \caption{Diffusion Noise Discretization}
    \label{fig:diffusion_rho}
  \end{subfigure}
  \hfill
  \begin{subfigure}{0.33\linewidth}
    \includegraphics[width=\linewidth]{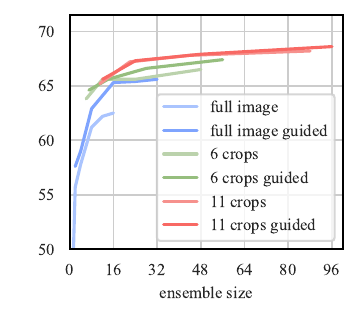}
    \caption{Test-time Ensembling}
    \label{fig:ensembling}
  \end{subfigure}
  \caption{\textbf{Ablation study on aspects of the diffusion process.} We study the impact of different parameters on the generative process: (a) the number of $N$ diffusion steps, (b) the result of different $\rho$ noise discretizations, and (c) the scaling ability of ensemble sizes.}
  \label{fig:short}
\vspace{-1.2em}
\end{figure*}

\subsection{Increasing Conditioning Objects at test-time}
A primary limitation of \acrshort{OC-DiT} is its resource demand, which increases with the number of conditioning objects. 
The primary reason is the attention layers in the OC-DiT blocks, as well as the overhead of loading numerous templates during training, which significantly slows convergence (\eg, training with 16 objects took $50\%$ longer than 8).
\begin{table}[b]
\vspace{-1.2em}
\centering
\begin{tabular}{|llc|}
\hline
\multicolumn{2}{|l}{Positional Encoding Method} & AP \\
\hline
\ref{item:pe_b1}& baseline 1: $2\times(8 \rightarrow 8)$ & 53.0  \\
\ref{item:pe_b2}& baseline 2: $16 \rightarrow 16$ & 60.3   \\
\ref{item:pe_tt_interp}& test-time interpolation: $8 \rightarrow 16$ & 55.6  \\
\ref{item:pe_shift} & random interval: $8 \rightarrow 16$ & \textbf{65.2}  \\
\ref{item:pe_interp}& random interpolate: $8 \rightarrow 16$ & 59.5   \\
\ref{item:pe_random} &random permutation: $8 \rightarrow 16$ & 58.7   \\
\hline
\end{tabular}
\caption{\textbf{Effects of positional encodings} to extrapolate to more objects during inference. The notation indicates the training and testing number of objects.}
\label{tab:pe_results}
\end{table}
To mitigate this, we investigate training with a smaller set of 8 objects and then scale to 16 during inference. 
We evaluate the following techniques for scaling object queries:
\begin{enumerate*}[label=(\roman*)]
    \item \label{item:pe_b1} \textit{Baseline 1}: Train on $N^\text{train}_\mathcal{O}$, repeat inference for all $N^\text{test}_\mathcal{O}$ objects.
    \item \label{item:pe_b2} \textit{Baseline 2}: Train directly on $N^\text{test}_\mathcal{O}$ objects.
    \item \label{item:pe_tt_interp} \textit{Test-time Interpolation}: Interpolate positional encodings from $N^\text{train}_\mathcal{O}$ to $N^\text{test}_\mathcal{O}$ at inference.
    \item \label{item:pe_shift} \textit{Random Interval}: Train with positional encoding capacity for $N^\text{test}_\mathcal{O}$, using random intervals of length $N^\text{train}_\mathcal{O}$.
    \item \label{item:pe_interp} \textit{Random Interpolation}: Train with positional encoding capacity for $N^\text{test}_\mathcal{O}$, using random intervals $> N^\text{train}_\mathcal{O}$ and interpolating to $N^\text{train}_\mathcal{O}$.
    \item \label{item:pe_random} \textit{Random Permutation}: Train with positional encoding capacity for $N^\text{test}_\mathcal{O}$, randomly select $N^\text{train}_\mathcal{O}$ indices.
\end{enumerate*}
See the supplementary materials for a visualization of the techniques.

\cref{tab:pe_results} shows the performance of the different encodings. \ref{item:pe_shift}  significantly outperforms all other approaches.
Most interestingly, \ref{item:pe_shift} even outperforms \ref{item:pe_b2}, which took around $50\%$ longer to train.
We hypothesize that this performance drop is due to the distribution of the objects per sample and the deteriorating effects on the training loss.

\subsection{Ablations}
\vspace{0.3em}\noindent\textbf{Diffusion Parameterization}.
\noindent This ablation examines the impact of diffusion inference parameters, specifically $\rho$ and $N$ sampling steps. 
Results on YCBV (\cref{fig:diffusion_steps}, \cref{fig:diffusion_rho}) show $N$ is crucial, with diminishing returns after the initial $9$ steps, despite linear inference time scaling. 
We investigate the effect of noise discretization $\rho$ (see \cref{eq:discretization}) for $N=9$ and $N=18$ steps. $\rho$ controls the noise level per step, with higher $\rho$ values emphasizing lower noise levels. 
Surprisingly, $\rho$ has only a small effect in our repeated experiments, consistently showing $\rho=9$ as marginally best.
In our experiments, the choice of $\rho$ is less important when diffusing latent codes for segmentation than when generating images \cite{karras2022elucidating}.
We hypothesize that potential reasons for this may be the log-uniform noise training distribution or the signal characteristics of segmentation data; however, further investigation is needed.

\vspace{0.3em}\noindent\textbf{Refinement.}
We evaluate the \texttt{refine} model's isolated performance and its discriminative capabilities.
\begin{table}
\centering
\begin{tabular}{|lcc|}
\hline
& YCBV & TUDL\\
\hline
\texttt{refine} w/o labels & 72.7 & 56.8  \\
\texttt{refine} w/ labels & 73.3 & 65.4  \\
\hline
\end{tabular}
\caption{\textbf{Ablation study on refinement.}  We evaluate the \texttt{refine} model's isolated performance and discriminative capabilities. w/ labels denotes conditioning on the true object, while w/o labels assesses discrimination of true/false positives from all objects.}
\label{tab:refinement_results}
\vspace{-1.2em}
\end{table}
We use the ground truth segmentations to generate modal crops for each object.
\begin{enumerate*}[label=(\alph*)]
\item \label{item:refine_0} \texttt{refine} w/ labels: for each crop, we only generate the segmentation conditioned on the true object.
\item  \label{item:refine_01} \texttt{refine} w/o labels: for each crop, we generate segmentations conditioned on all possible individual objects, such that the model has to discriminate true positives from false positives. 
\end{enumerate*}
The latter simulates a common use-case where open-set detectors generate unlabeled proposals that need to be assigned a class label~\cite{nguyen2023cnos, liu2024grounding, kirillov2023segment}.
We present the results in \cref{tab:refinement_results}.
In \ref{item:refine_0} \texttt{refine} achieves a  $65.4$ \acrshort{AP} on the TUDL dataset. 
On this dataset, the best fully-supervised method achieves $\sim  75.0$~\acrshort{AP}.
On YCBV data, our model achieves $73.3$ \acrshort{AP} on par with the best supervised 2D segmentation results.
In \ref{item:refine_01}, our model achieves an \acrshort{AP} of $56.8$ for TUDL and a very strong $72.7$ for YBC, demonstrating the discriminative power of the model.

\vspace{0.3em}\noindent\textbf{Training Data.}
To evaluate the impact of training data, we train identical models on subsets of data introduced in \cref{subsec:implementation}.
As shown in Table~\ref{tab:datasets_results}, both new datasets significantly improve the segmentation performance evaluated on YCBV. 
Interestingly, the objaverse dataset has the highest positive impact.
Nonetheless, the samples from MegaPoseGSO have a significant contribution.
A potential reason could be insufficient training objects in \acrshort{GSO}, or that the objaverse dataset may contain more objects similar to the YCBV test objects. 
Using all datasets performs best, suggesting that scaling the method further is possible.
\begin{table}[b]
\vspace{-1.2em}
\centering
\begin{tabular}{|llc|}
\hline
\multicolumn{2}{|l}{Training Data Combinations }& \acrshort{AP} \\
\hline
\ref{item:megapose} &MegaPose \acrshort{GSO} & 56.3  \\
\ref{item:obd_gso} &Our \acrshort{GSO} & 60.3  \\
\ref{item:obd_objaverse} &Our Objaverse & 62.0   \\
\ref{item:megapose}, \ref{item:obd_objaverse} & MegaPose GSO + Our Objaverse & 63.1  \\
\ref{item:obd_gso},\ref{item:obd_objaverse} &Our GSO + Our Objaverse & 62.0  \\
\ref{item:megapose}, \ref{item:obd_gso}, \ref{item:obd_objaverse} &All three datasets & 65.5 \\
\hline
\end{tabular}
\caption{\textbf{Ablation study of training datasets.} We study the impact of combinations of training data on the \acrshort{AP} on YCBV.}
\label{tab:datasets_results}
\end{table}

\vspace{0.3em}\noindent\textbf{Test-time Ensembling.}
We assess test-time ensembling with three schemes: 'full image', '6 crops' (full image + 5 spatial augmentations), and '11 crops' (full image + 10 spatial augmentations). 
We visualize the spatial augmentations in the supplementary materials. 
An additional guided augmentation, tightly bounding previously estimated objects, is also evaluated. 
\cref{fig:ensembling} shows that increasing ensemble size ($N_\text{ensemble} \times N_\text{augs}$) drastically improves performance.

\subsection{Limitations}
While \acrshort{OC-DiT} excels in complex scene analysis, its slow inference hinders real-time applications. 
Diffusion models' iterative sampling linearly increases inference time with steps and ensemble size, significantly impacting speed (see \cref{fig:diffusion_steps}, \cref{fig:ensembling}, \cref{tab:inference_time}). 
Furthermore, performance heavily relies on training data (\cref{tab:datasets_results}), requiring further analysis to understand the influence of specific data characteristics. 
Available large-scale datasets with sufficient object variety and instance counts are lacking. 
Finally, scaling \acrshort{OC-DiT} to a large number of objects is computationally expensive due to increased query and feature processing.
\begin{table}
\centering
\begin{tabular}{|ll|cccc|}
\hline
\multirow{2}{*}{Resolution} & \multirow{2}{*}{$N_\mathcal{O}$} & \multicolumn{4}{c|}{Ensemble Size} \\
\cline{3-6}
& & 1 & 4 & 16 & 64 \\
\hline
$252 \times 336$ &8 & 2.8 \unit{\second} &  5.6 \unit{\second}& 22.2 \unit{\second}& 88.9 \unit{\second} \\
$252 \times 336$ &16 & 5.0 \unit{\second}& 14.5 \unit{\second}& 58.1 \unit{\second}& 232.4 \unit{\second}\\
$252 \times 336$ &24 & 9.1 \unit{\second}& 26.7 \unit{\second}& 106.8 \unit{\second}& 403.2 \unit{\second}\\

$420 \times 560$ &8 & 6.8 \unit{\second}& 21.5 \unit{\second}& 85.3 \unit{\second}& 341.6 \unit{\second}\\

\hline
\end{tabular}
\caption{\textbf{Study on inference time.} Processing time scales with input resolution, object conditions, and ensemble size.}
\label{tab:inference_time}
\vspace{-1.2em}
\end{table}
\section{Conclusion}
\label{sec:conclusion}
We presented \acrfull{OC-DiT}, a novel formulation to instance segmentation based on a conditional generative diffusion process, and a state-of-the-art zero-shot instance segmentation model.
We rigorously tested \acrshort{OC-DiT} on a wide range of objects in cluttered scenes. 
We investigated methods to increase the number of query objects at test-time, reducing training time and improving inference performance.
In addition, we conducted thorough ablation studies to validate our design choices and demonstrate the critical role of training data.
We will release our extensive synthetic datasets and code to encourage further research in zero-shot estimation. 
Future work will focus on reducing memory requirements, improving efficiency for more query objects, and tailoring the diffusion framework specifically to instance segmentation.
{
    \small
    \bibliographystyle{ieeenat_fullname}
    \bibliography{main}

\begin{thebibliography}{50}
\providecommand{\natexlab}[1]{#1}
\providecommand{\url}[1]{\texttt{#1}}
\expandafter\ifx\csname urlstyle\endcsname\relax
  \providecommand{\doi}[1]{doi: #1}\else
  \providecommand{\doi}{doi: \begingroup \urlstyle{rm}\Url}\fi

\bibitem[Amit et~al.(2021)Amit, Shaharbany, Nachmani, and Wolf]{amit2021segdiff}
T. Amit, T. Shaharbany, E. Nachmani, and L. Wolf.
\newblock Segdiff: Image segmentation with diffusion probabilistic models.
\newblock \emph{arXiv preprint arXiv:2112.00390}, 2021.

\bibitem[Brachmann et~al.(2014)Brachmann, Krull, Michel, Gumhold, Shotton, and Rother]{brachmann2014learning}
E. Brachmann, A. Krull, F. Michel, S. Gumhold, J. Shotton, and C. Rother.
\newblock Learning 6d object pose estimation using 3d object coordinates.
\newblock In \emph{ECCV}, pages 536--551. Springer, 2014.

\bibitem[Carion et~al.(2020)Carion, Massa, Synnaeve, Usunier, Kirillov, and Zagoruyko]{carion_end_2020}
N. Carion, F. Massa, G. Synnaeve, N. Usunier, A. Kirillov, and S. Zagoruyko.
\newblock End-to-end object detection with transformers.
\newblock In \emph{ECCV}, pages 213--229. Springer International Publishing, 2020.

\bibitem[Chen et~al.(2023)Chen, Sun, Bao, Zhao, Wu, and He]{chen2023zeropose}
Jianqiu Chen, Mingshan Sun, Tianpeng Bao, Rui Zhao, Liwei Wu, and Zhenyu He.
\newblock Zeropose: Cad-model-based zero-shot pose estimation.
\newblock \emph{arXiv preprint arXiv:2305.17934}, 2023.

\bibitem[Deitke et~al.(2023)Deitke, Schwenk, Salvador, Weihs, Michel, VanderBilt, Schmidt, Ehsani, Kembhavi, and Farhadi]{deitke2023objaverse}
M. Deitke, D. Schwenk, J. Salvador, L. Weihs, O. Michel, E. VanderBilt, L. Schmidt, K. Ehsani, A. Kembhavi, and A. Farhadi.
\newblock Objaverse: A universe of annotated 3d objects.
\newblock In \emph{CVPR}, pages 13142--13153, 2023.

\bibitem[Denninger et~al.(2019)Denninger, Sundermeyer, Winkelbauer, Zidan, Olefir, Elbadrawy, Lodhi, and Katam]{denninger2019blenderproc}
M. Denninger, M. Sundermeyer, D. Winkelbauer, Y. Zidan, D. Olefir, M. Elbadrawy, A. Lodhi, and H. Katam.
\newblock Blenderproc.
\newblock \emph{arXiv preprint arXiv:1911.01911}, 2019.

\bibitem[Dhariwal and Nichol(2021)]{dhariwal2021diffusion}
P. Dhariwal and A. Nichol.
\newblock Diffusion models beat gans on image synthesis.
\newblock \emph{Advances in neural information processing systems}, 34:\penalty0 8780--8794, 2021.

\bibitem[Dosovitskiy(2020)]{dosovitskiy2020image}
A. Dosovitskiy.
\newblock An image is worth 16x16 words: Transformers for image recognition at scale.
\newblock \emph{arXiv preprint arXiv:2010.11929}, 2020.

\bibitem[Downs et~al.(2022)Downs, Francis, Koenig, Kinman, Hickman, Reymann, McHugh, and Vanhoucke]{downs2022google}
L. Downs, A. Francis, N. Koenig, B. Kinman, R. Hickman, K. Reymann, T. McHugh, and V. Vanhoucke.
\newblock Google scanned objects: A high-quality dataset of 3d scanned household items.
\newblock In \emph{ICRA}, pages 2553--2560, 2022.

\bibitem[Du et~al.(2021)Du, Xiao, and Lepetit]{du2021learning}
Y. Du, Y. Xiao, and V. Lepetit.
\newblock Learning to better segment objects from unseen classes with unlabeled videos.
\newblock In \emph{ICCV}, pages 3375--3384, 2021.

\bibitem[Durner et~al.(2021)Durner, Boerdijk, Sundermeyer, Friedl, M{\'a}rton, and Triebel]{durner2021unknown}
M. Durner, W. Boerdijk, M. Sundermeyer, W. Friedl, Z.-C. M{\'a}rton, and R. Triebel.
\newblock Unknown object segmentation from stereo images.
\newblock In \emph{IROS}, pages 4823--4830. IEEE, 2021.

\bibitem[He et~al.(2017)He, Gkioxari, Dollár, and Girshick]{he_mask_2017}
K. He, G. Gkioxari, P. Dollár, and R. Girshick.
\newblock Mask {R}-{CNN}.
\newblock In \emph{ICCV}, pages 2980--2988, 2017.

\bibitem[Hendrycks and Gimpel(2016)]{hendrycks2016gaussian}
D. Hendrycks and K. Gimpel.
\newblock Gaussian error linear units (gelus).
\newblock \emph{arXiv preprint arXiv:1606.08415}, 2016.

\bibitem[Ho et~al.(2020)Ho, Jain, and Abbeel]{ho2020denoising}
J. Ho, A. Jain, and P. Abbeel.
\newblock Denoising diffusion probabilistic models.
\newblock \emph{NeurIPS}, 33:\penalty0 6840--6851, 2020.

\bibitem[Hodaň et~al.(2024)Hodaň, Sundermeyer, Labbé, Nguyen, Wang, Brachmann, Drost, Lepetit, Rother, and Matas]{hodan_bop_2024}
T. Hodaň, M. Sundermeyer, Y. Labbé, V.~N. Nguyen, G. Wang, E. Brachmann, B. Drost, V. Lepetit, C. Rother, and J. Matas.
\newblock {BOP} {Challenge} 2023 on {Detection}, {Segmentation} and {Pose} {Estimation} of {Seen} and {Unseen} {Rigid} {Objects}.
\newblock \emph{CVPRW}, 2024.

\bibitem[Humt et~al.(2023)Humt, Winkelbauer, and Hillenbrand]{humt_shape_2023}
M. Humt, D. Winkelbauer, and U. Hillenbrand.
\newblock Shape {Completion} with {Prediction} of {Uncertain} {Regions}.
\newblock In \emph{IROS}, 2023.

\bibitem[Hyv{\"a}rinen and Dayan(2005)]{hyvarinen2005estimation}
A. Hyv{\"a}rinen and P. Dayan.
\newblock Estimation of non-normalized statistical models by score matching.
\newblock \emph{JMLR}, 6\penalty0 (4), 2005.

\bibitem[Karras et~al.(2022)Karras, Aittala, Aila, and Laine]{karras2022elucidating}
T. Karras, M. Aittala, T. Aila, and S. Laine.
\newblock Elucidating the design space of diffusion-based generative models.
\newblock \emph{NeurIPS}, 35:\penalty0 26565--26577, 2022.

\bibitem[Karras et~al.(2024)Karras, Aittala, Lehtinen, Hellsten, Aila, and Laine]{karras2024analyzing}
T. Karras, M. Aittala, J. Lehtinen, J. Hellsten, T. Aila, and S. Laine.
\newblock Analyzing and improving the training dynamics of diffusion models.
\newblock In \emph{CVPR}, pages 24174--24184, 2024.

\bibitem[Kaskman et~al.(2019)Kaskman, Zakharov, Shugurov, and Ilic]{kaskman2019homebreweddb}
R. Kaskman, S. Zakharov, I. Shugurov, and S. Ilic.
\newblock Homebreweddb: Rgb-d dataset for 6d pose estimation of 3d objects.
\newblock In \emph{ICCVW}, pages 0--0, 2019.

\bibitem[Kendall et~al.(2018)Kendall, Gal, and Cipolla]{kendall2018multi}
A. Kendall, Y. Gal, and R. Cipolla.
\newblock Multi-task learning using uncertainty to weigh losses for scene geometry and semantics.
\newblock In \emph{CVPR}, pages 7482--7491, 2018.

\bibitem[Kirillov et~al.(2023)Kirillov, Mintun, Ravi, Mao, Rolland, Gustafson, Xiao, Whitehead, Berg, Lo, et~al.]{kirillov2023segment}
A. Kirillov, E. Mintun, N. Ravi, H. Mao, C. Rolland, L. Gustafson, T. Xiao, S. Whitehead, A. Berg, W.-Y. Lo, et~al.
\newblock Segment anything.
\newblock In \emph{ICCV}, pages 4015--4026, 2023.

\bibitem[Labb{\'e} et~al.(2022)Labb{\'e}, Manuelli, Mousavian, Tyree, Birchfield, Tremblay, Carpentier, Aubry, Fox, and Sivic]{labbe2022megapose}
Y. Labb{\'e}, L. Manuelli, A. Mousavian, S. Tyree, S. Birchfield, J. Tremblay, J. Carpentier, M. Aubry, D. Fox, and J. Sivic.
\newblock Megapose: 6d pose estimation of novel objects via render \& compare.
\newblock \emph{arXiv preprint arXiv:2212.06870}, 2022.

\bibitem[Lin et~al.(2024{\natexlab{a}})Lin, Liu, Lu, and Jia]{lin2024sam}
J. Lin, L. Liu, D. Lu, and K. Jia.
\newblock Sam-6d: Segment anything model meets zero-shot 6d object pose estimation.
\newblock In \emph{CVPR}, pages 27906--27916, 2024{\natexlab{a}}.

\bibitem[Lin et~al.(2024{\natexlab{b}})Lin, Su, Nathan, Inuganti, Di, Sundermeyer, Manhardt, Stricker, Rambach, and Zhang]{lin_hipose_2024}
Y. Lin, Y. Su, P. Nathan, S. Inuganti, Y. Di, M. Sundermeyer, F. Manhardt, D. Stricker, J. Rambach, and Y. Zhang.
\newblock Hipose: Hierarchical binary surface encoding and correspondence pruning for rgb-d 6dof object pose estimation.
\newblock In \emph{CVPR}, pages 10148--10158, 2024{\natexlab{b}}.

\bibitem[Liu et~al.(2024)Liu, Zeng, Ren, Li, Zhang, Yang, Jiang, Li, Yang, Su, et~al.]{liu2024grounding}
Shilong Liu, Zhaoyang Zeng, Tianhe Ren, Feng Li, Hao Zhang, Jie Yang, Qing Jiang, Chunyuan Li, Jianwei Yang, Hang Su, et~al.
\newblock Grounding dino: Marrying dino with grounded pre-training for open-set object detection.
\newblock In \emph{European Conference on Computer Vision}, pages 38--55. Springer, 2024.

\bibitem[Lu et~al.(2024)Lu, Guo, Ruozzi, Xiang, et~al.]{lu2024adapting}
Yangxiao Lu, Yunhui Guo, Nicholas Ruozzi, Yu Xiang, et~al.
\newblock Adapting pre-trained vision models for novel instance detection and segmentation.
\newblock \emph{arXiv preprint arXiv:2405.17859}, 2024.

\bibitem[Newbury et~al.(2023)Newbury, Gu, Chumbley, Mousavian, Eppner, Leitner, Bohg, Morales, Asfour, Kragic, et~al.]{newbury2023deep}
R. Newbury, M. Gu, L. Chumbley, A. Mousavian, C. Eppner, J. Leitner, J. Bohg, A. Morales, T. Asfour, D. Kragic, et~al.
\newblock Deep learning approaches to grasp synthesis: A review.
\newblock \emph{IEEE Transactions on Robotics}, 39\penalty0 (5):\penalty0 3994--4015, 2023.

\bibitem[Nguyen et~al.(2023)Nguyen, Groueix, Ponimatkin, Lepetit, and Hodan]{nguyen2023cnos}
V.~N. Nguyen, T. Groueix, G. Ponimatkin, V. Lepetit, and T. Hodan.
\newblock Cnos: A strong baseline for cad-based novel object segmentation.
\newblock In \emph{ICCV}, pages 2134--2140, 2023.

\bibitem[Nichol and Dhariwal(2021)]{nichol2021improved}
A.~Q. Nichol and P. Dhariwal.
\newblock Improved denoising diffusion probabilistic models.
\newblock In \emph{ICML}, pages 8162--8171. PMLR, 2021.

\bibitem[Oquab et~al.(2023)Oquab, Darcet, Moutakanni, Vo, Szafraniec, Khalidov, Fernandez, Haziza, Massa, El-Nouby, et~al.]{oquab2023dinov2}
M. Oquab, T. Darcet, T. Moutakanni, H. Vo, M. Szafraniec, V. Khalidov, P. Fernandez, D. Haziza, F. Massa, A. El-Nouby, et~al.
\newblock Dinov2: Learning robust visual features without supervision.
\newblock \emph{arXiv preprint arXiv:2304.07193}, 2023.

\bibitem[Peebles and Xie(2023)]{peebles2023scalable}
W. Peebles and S. Xie.
\newblock Scalable diffusion models with transformers.
\newblock In \emph{ICCV}, pages 4195--4205, 2023.

\bibitem[Perez et~al.(2018)Perez, Strub, De~Vries, Dumoulin, and Courville]{perez2018film}
E. Perez, F. Strub, H. De~Vries, V. Dumoulin, and A. Courville.
\newblock Film: Visual reasoning with a general conditioning layer.
\newblock In \emph{AAAI}, 2018.

\bibitem[Rahman et~al.(2023)Rahman, Valanarasu, Hacihaliloglu, and Patel]{rahman2023ambiguous}
A. Rahman, J.~M.~J. Valanarasu, I. Hacihaliloglu, and V.~M. Patel.
\newblock Ambiguous medical image segmentation using diffusion models.
\newblock In \emph{CVPR}, pages 11536--11546, 2023.

\bibitem[Rombach et~al.(2022)Rombach, Blattmann, Lorenz, Esser, and Ommer]{rombach2022high}
R. Rombach, A. Blattmann, D. Lorenz, P. Esser, and B. Ommer.
\newblock High-resolution image synthesis with latent diffusion models.
\newblock In \emph{CVPR}, pages 10684--10695, 2022.

\bibitem[Sohl-Dickstein et~al.(2015)Sohl-Dickstein, Weiss, Maheswaranathan, and Ganguli]{Sohl-Dickstein2015-qp}
J. Sohl-Dickstein, E. Weiss, N. Maheswaranathan, and S. Ganguli.
\newblock Deep unsupervised learning using nonequilibrium thermodynamics.
\newblock \emph{arXiv [cs.LG]}, 2015.

\bibitem[Song et~al.(2020{\natexlab{a}})Song, Meng, and Ermon]{song2020denoising}
J. Song, C. Meng, and S. Ermon.
\newblock Denoising diffusion implicit models.
\newblock \emph{arXiv preprint arXiv:2010.02502}, 2020{\natexlab{a}}.

\bibitem[Song et~al.(2020{\natexlab{b}})Song, Sohl-Dickstein, Kingma, Kumar, Ermon, and Poole]{song2020score}
Y. Song, J. Sohl-Dickstein, D.~P. Kingma, A. Kumar, S. Ermon, and B. Poole.
\newblock Score-based generative modeling through stochastic differential equations.
\newblock \emph{arXiv preprint arXiv:2011.13456}, 2020{\natexlab{b}}.

\bibitem[Stoiber et~al.(2022)Stoiber, Sundermeyer, and Triebel]{stoiber_iterative_2022}
M. Stoiber, M. Sundermeyer, and R. Triebel.
\newblock Iterative {Corresponding} {Geometry}: {Fusing} {Region} and {Depth} for {Highly} {Efficient} {3D} {Tracking} of {Textureless} {Objects}.
\newblock In \emph{CVPR}. IEEE, 2022.

\bibitem[Sundermeyer et~al.(2021)Sundermeyer, Mousavian, Triebel, and Fox]{sundermeyer_contact_2021}
M. Sundermeyer, A. Mousavian, R. Triebel, and D. Fox.
\newblock Contact-graspnet: Efficient 6-dof grasp generation in cluttered scenes.
\newblock In \emph{ICRA}, pages 13438--13444. IEEE, 2021.

\bibitem[Sundermeyer et~al.(2023)Sundermeyer, Hoda{\v{n}}, Labbe, Wang, Brachmann, Drost, Rother, and Matas]{sundermeyer2023bop}
M. Sundermeyer, T. Hoda{\v{n}}, Y. Labbe, G. Wang, E. Brachmann, B. Drost, C. Rother, and J. Matas.
\newblock {BOP} {Challenge} 2023 on {Detection}, {Segmentation} and {Pose} {Estimation} of {Specific} {Rigid} {Objects}.
\newblock In \emph{CVPRW}, pages 2785--2794, 2023.

\bibitem[Tan et~al.(2023)Tan, Chen, and Yan]{tan2023diffss}
W. Tan, S. Chen, and B. Yan.
\newblock Diffss: Diffusion model for few-shot semantic segmentation.
\newblock \emph{arXiv preprint arXiv:2307.00773}, 2023.

\bibitem[Ulmer et~al.(2023)Ulmer, Durner, Sundermeyer, Stoiber, and Triebel]{ulmer20236d}
M. Ulmer, M. Durner, M. Sundermeyer, M. Stoiber, and R. Triebel.
\newblock 6d object pose estimation from approximate 3d models for orbital robotics.
\newblock In \emph{IROS}, pages 10749--10756. IEEE, 2023.

\bibitem[Wolleb et~al.(2022)Wolleb, Sandk{\"u}hler, Bieder, Valmaggia, and Cattin]{wolleb2022diffusion}
J. Wolleb, R. Sandk{\"u}hler, F. Bieder, P. Valmaggia, and P.~C. Cattin.
\newblock Diffusion models for implicit image segmentation ensembles.
\newblock In \emph{MIDL}, pages 1336--1348. PMLR, 2022.

\bibitem[Xiang et~al.(2017)Xiang, Schmidt, Narayanan, and Fox]{xiang2017posecnn}
Y. Xiang, T. Schmidt, V. Narayanan, and D. Fox.
\newblock Posecnn: A convolutional neural network for 6d object pose estimation in cluttered scenes.
\newblock \emph{arXiv preprint arXiv:1711.00199}, 2017.

\bibitem[Xiang et~al.(2021)Xiang, Xie, Mousavian, and Fox]{xiang2021learning}
Y. Xiang, C. Xie, A. Mousavian, and D. Fox.
\newblock Learning rgb-d feature embeddings for unseen object instance segmentation.
\newblock In \emph{Conference on Robot Learning}, pages 461--470. PMLR, 2021.

\bibitem[Xie et~al.(2021)Xie, Xiang, Mousavian, and Fox]{xie2021unseen}
C. Xie, Y. Xiang, A. Mousavian, and D. Fox.
\newblock Unseen object instance segmentation for robotic environments.
\newblock \emph{IEEE Transactions on Robotics}, 37\penalty0 (5):\penalty0 1343--1359, 2021.

\bibitem[Yan et~al.(2022)Yan, Lin, Mitra, Lischinski, Cohen-Or, and Huang]{yan_shapeformer_2022}
X. Yan, L. Lin, N.~J. Mitra, D. Lischinski, D. Cohen-Or, and H. Huang.
\newblock {ShapeFormer}: {Transformer}-based {Shape} {Completion} via {Sparse} {Representation}.
\newblock In \emph{CVPR}, pages 6229--6239, 2022.

\bibitem[Z. et~al.(2023)Z., D., A., D., Y., L., T., and W.]{zhao2023fast}
Xu Z., Wenchao D., Yongqi A., Yinglong D., Tao Y., Min L., Ming T., and Jinqiao W.
\newblock Fast segment anything, 2023.

\bibitem[Zheng et~al.(2021)Zheng, Wu, Qin, Zhang, and Cui]{zheng2021zero}
Y. Zheng, J. Wu, Y. Qin, F. Zhang, and L. Cui.
\newblock Zero-shot instance segmentation.
\newblock In \emph{CVPR}, pages 2593--2602, 2021.

\end{thebibliography}
}
\clearpage
\appendix
\section{Implementation and Training Details}
\subsection{Object Templates Rendering}
\begin{figure*}
  \centering
\includegraphics[width=\linewidth]{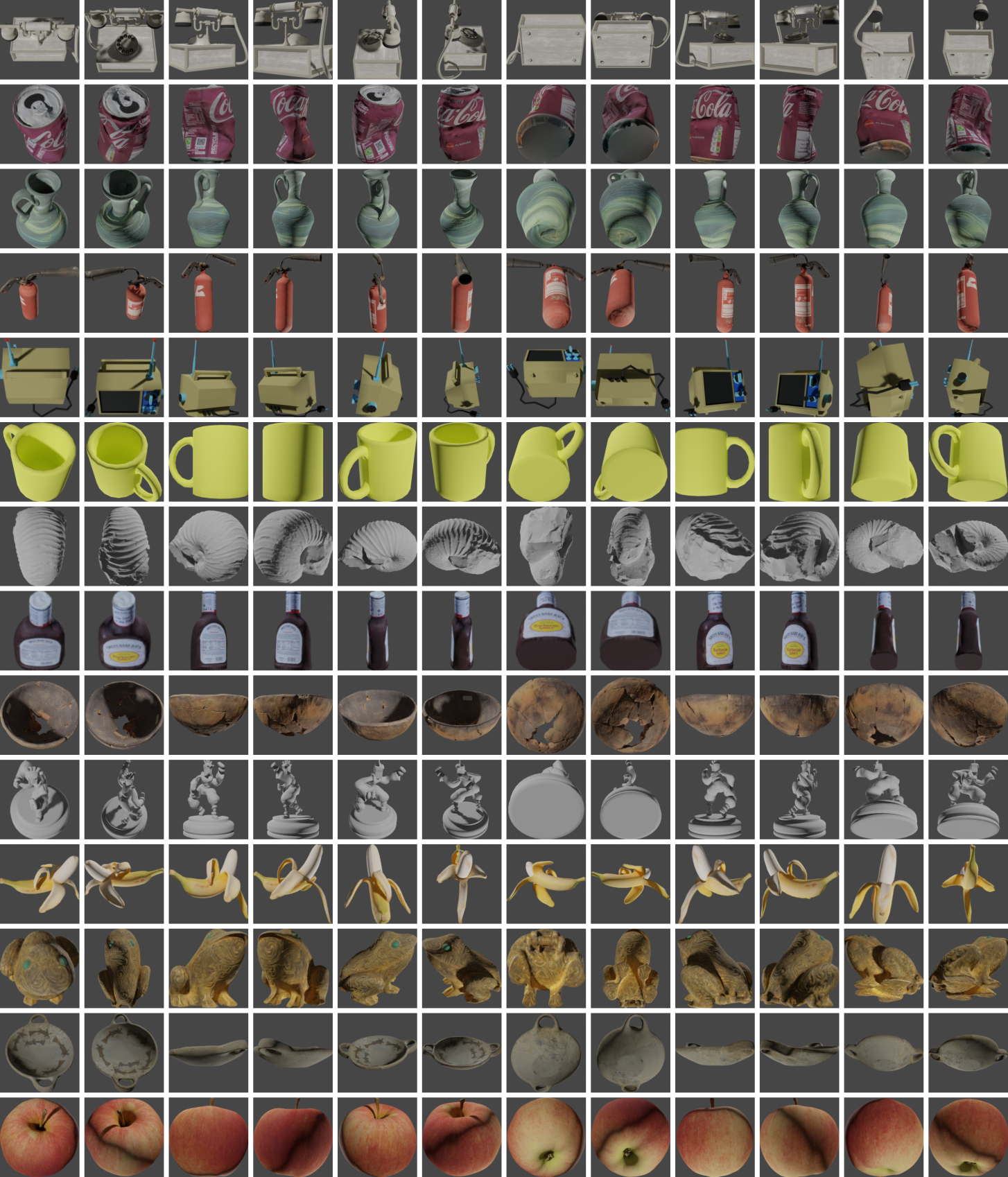}
\caption{\textbf{Templates rendered for objaverse~\cite{deitke2023objaverse} objects.}}
\label{fig:templates}
\end{figure*}

We use BlenderProc\cite{denninger2019blenderproc} for template generation and use the camera intrinsics from the TUDL dataset.
We scale each model and render $42$ viewpoints distributed on an icosphere. 
The radius of this viewsphere is dynamically determined based on the size of the object, ensuring it remains within the camera's field of view. 
We add directional lighting in such a way that there are shadows that show the geometry of the model.
For each rendered view, we generate RGB images and object masks, crop and resizes them to a standardized resolution of $224 \times 224$, and calculates bounding boxes around the object. 
The rendered images, masks, bounding boxes, 3D point clouds of the object, camera intrinsics, and camera poses are then saved into a compressed NumPy archive for each model.
Please refer to \Cref{fig:templates} for a visualization of templates rendered for objaverse~\cite{deitke2023objaverse} objects.

\subsection{Variational Autoencoder (VAE) Training Details}
We train the \acrshort{VAE} on binary masks from the same training dataset as we train \acrshort{OC-DiT} and tune the hyperparameters, such that the latent has zero mean and standard deviation $1.0$. 
The training procedure focuses on learning a robust latent representation of object masks using a convolutional variational autoencoder with attention mechanisms. 
Data is loaded according to the dataset in \ref{subsec:implementation}. However, we load all templates and generate masks to increase the number of training samples per disk operation.
Extensive data augmentation is performed using torchvision's transforms.v2.
This includes random horizontal and vertical flips, random affine transformations (rotation, translation, scaling), and random perspective transformations.
These augmentations aim to improve the model's generalization capabilities.
The latent space has a dimensionality of 64 channels, and the model processes single-channel mask images as both input and output.
During the forward pass, augmented mask images are fed into the encoder of the VAE. 
The encoder produces a latent representation, from which the decoder reconstructs the mask.
The training process utilizes the Evidence Lower Bound (ELBO) loss, with a beta parameter of $10^{-5}$ to control the KL divergence term. 
This balances reconstruction accuracy and latent space regularization.
Optimization is performed using the Adam optimizer with a learning rate of $3 \times 10^{-4}$ and no weight decay. A learning rate scheduler is employed, featuring a linear warmup followed by cosine annealing.
The training proceeds in batches of size 128 for 70 epochs, with the learning rate adjusted by the scheduler.

\subsection{OC-DiT Training Details and Hyperparameters}
For each sample of the dataset that we load, we have to reduce the number of objects per sample to the desired number.
We only load the templates for the objects that we select in the following process.
We sample a random point in the image and calculate a weighting scalar dependent on the distance to the sampled point. 
We use this weighting to randomize the different objects that we keep in the sample.
This allows us to bias the sampling process to choose objects that are closer to one another, resulting in better training signal.
\acrshort{OC-DiT} leverages a pre-trained variational autoencoder (VAE) for latent space operations that has to be pre-trained.

\noindent\texttt{coarse} is trained on $252\times336$ images and $288\times384$ segmentations ($18\times24$ patches), conditioned on multiple objects per sample. 
The decoder used $10$ blocks with a hidden size $1152$, and $12$ attention heads. 
Training lasted 150 epochs, with $1000$ steps per epoch and batch size $16$ on $2$ A100 GPUs, using the Adam optimizer with learning rate of $2\cdot 10^{-4}$), linear warmup for 2000 steps, and cosine annealing after $72\%$).
Data augmentation included random crops and jittered bounding boxes around all conditioning objects.

\noindent\texttt{refine} is trained on $224 \times 224$ images and $256 \times 256$ segmentations ($16 \times 16$ patches), conditioned on single objects. 
The decoder uses $10$ blocks, with a hidden size $1152$, and $12$ attention heads. 
Training lasted $200$ epochs, with $1000$ steps per epoch, and batch size $128$ on $2$ A100 GPUs, using the Adam optimizer with learning rate $1\cdot 10^{-4}$, linear warmup for $2000$ steps, and cosine annealing after $72\%$. 
Training data consists of cropped, jittered object regions. 

In this work we use $12$ template views due to VRAM constraints, but any number of templates can be used.
We train the model on $8$ object classes per sample which is a tradeoff between number of objects per sample and computational overhead.
The models object capacity is $16$. 
We randomly crop the image with a probability of  $0.7$ with a scale of $[0.5, 1.1]$ meaning that the image is randomly cropped to a size by this scale.
If we do not randomly crop the image, we perform a tight crop around all object instances that we jitter.
The Adam optimizer is employed with a learning rate of $2e-4$ and no weight decay. 
We use a linear warmup for the learning rate for $2000$ gradient steps and a cosine annealing after $0.72\%$ of the training is complete.
The training uses a batch size of $8$ per GPU. 
We apply two types of data augmentations, RGB augmentations with a probability of 0.9 and background augmentations with a probability of 0.2. 
The RGB augmentations entail blur, sharpness, contrast, brightness and a basic color augmentation.
We use an iterative dataset and each epoch is complete after  $1000$ training batche with a maximum of $120$ epochs. 
Please refer to \Cref{fig:examples_0} and \Cref{fig:examples_1} for visualization of the diffusion process of a trained model.

\subsection{Positional Encodings}
\begin{figure*}
  \centering
\includegraphics[width=\linewidth]{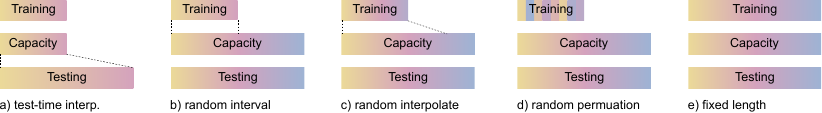}
\caption{\textbf{Positional encodings and adaption methods ablated in this paper.} a) we train a model with a fixed capacity and interpolate the positional embedding during inference to the desired size. b) we train a model with a higher capacity than the number of training objects and randomly select intervals during training. c) same as b) but we sample intervals that are larger than the training number of objects and interpolate the positional encoding to number of training objects d) we train the model with a higher capacity and randomly select positional codes e) we train the model directly on the increased number of objects.}
\label{fig:pe}
\end{figure*}

We evaluate the following techniques for scaling object queries, addressing the critical challenge of adapting models trained on a specific number of objects $N^\text{train}_\mathcal{O}$ to handle a potentially different number of objects at inference ($N^\text{test}_\mathcal{O}$). We delve into various approaches, each visualized in \Cref{fig:pe}.
\begin{enumerate}[label=(\roman*)]
    \item \textit{Baseline 1: Sequential Inference after Training on Limited Objects.} The model is trained on $N^\text{train}_\mathcal{O}$ objects. Subsequently, during inference, the model processes each of the $N^\text{test}_\mathcal{O}$ object chunks individually, effectively performing a series of independent inferences. This method tests the model's ability to generalize to unseen objects within the fixed training capacity without any explicit scaling mechanisms. A clear downside is that during self-attention, not all queries can attend to one another. This leads to scenarios where we can get multiple for one segment of the image.

    \item \textit{Baseline 2: Direct Training with Target Object Count.} As another baseline, this method directly trains the model on a dataset that precisely matches the number of objects anticipated at inference, $N^\text{test}_\mathcal{O}$. This provides a performance ceiling, showcasing the optimal results achievable when the training and inference object counts are perfectly aligned. It allows us to assess the performance loss when object counts are not aligned. However, as we see in the results, due to training dynamics this baseline performs worse than other positional encoding methods.

    \item \textit{Test-Time Positional Encoding Interpolation: Adapting to Varying Object Counts at Inference.} This technique addresses the mismatch between $N^\text{train}_\mathcal{O}$ and $N^\text{test}_\mathcal{O}$ by dynamically adjusting the positional encodings during the inference phase. Specifically, the positional encodings learned during training for $N^\text{train}_\mathcal{O}$ objects are interpolated to accommodate the $N^\text{test}_\mathcal{O}$ objects. This allows the model to handle a different number of objects without requiring retraining, offering flexibility in deployment scenarios.

    \item \textit{Random Interval Training: Enhancing Robustness through Variable Subsets.} This method trains the model with a positional encoding capacity sufficient to handle the maximum number of objects, $N^\text{test}_\mathcal{O}$. However, during each training iteration, only a random, contiguous interval of length $N^\text{train}_\mathcal{O}$ is utilized. This approach aims to improve the model's robustness and generalization by exposing it to various subsets of the larger positional encoding space, preventing overfitting to a specific object ordering.

    \item \textit{Random Interpolation Training: Leveraging Interpolation for Flexible Object Handling.} Similar to the previous technique, this method trains the model with a positional encoding capacity designed for $N^\text{test}_\mathcal{O}$ objects. However, instead of using contiguous intervals, random intervals of lengths greater than $N^\text{train}_\mathcal{O}$ are selected. Subsequently, the positional encodings within these intervals are interpolated down to $N^\text{train}_\mathcal{O}$ during training. 

    \item \textit{Random Permutation Training: Promoting Generalization through Diverse Object Orderings.} This method also leverages a positional encoding capacity designed for $N^\text{test}_\mathcal{O}$ objects. During each training iteration, a random subset of $N^\text{train}_\mathcal{O}$ indices is selected from the full range of $N^\text{test}_\mathcal{O}$ indices, effectively permuting the object order. This random permutation of indices aims to improve the model's generalization by exposing it to diverse orderings of the input objects, mitigating biases related to specific object arrangements.
\end{enumerate}

\subsection{Common Failure Modes of \texttt{coarse}}
\begin{figure*}
  \centering
\includegraphics[width=\linewidth]{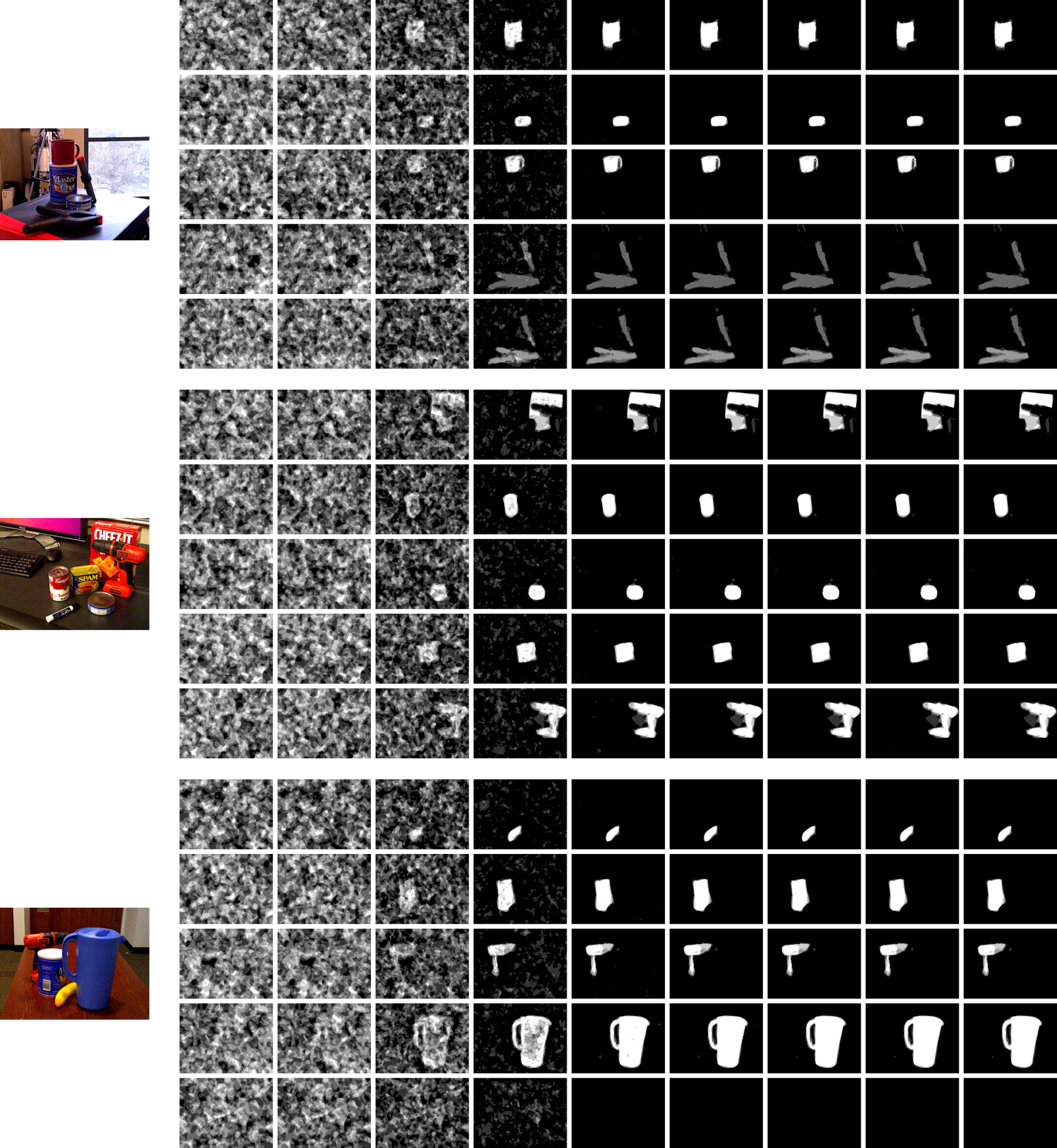}
\caption{Some examples of the generative process using $\rho=15$ for noise discretization.}
\label{fig:examples_0}
\end{figure*}

\begin{figure*}
  \centering
\includegraphics[width=\linewidth]{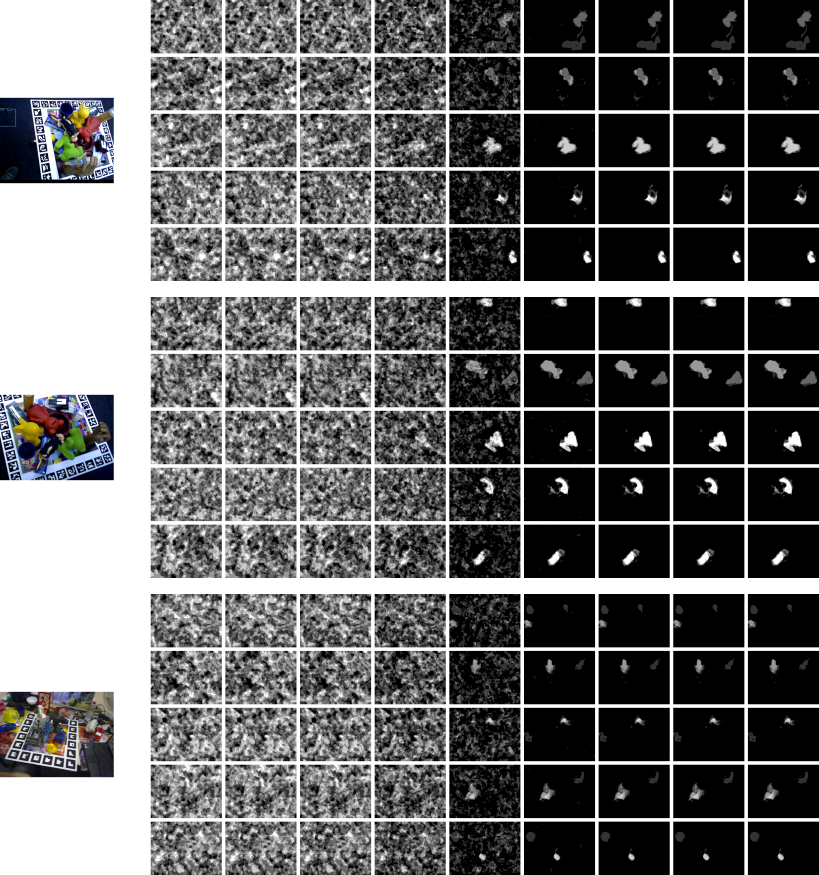}
\caption{Some examples of the generative process using $\rho=5$ for noise discretization.}
\label{fig:examples_1}
\end{figure*}

In \Cref{fig:examples_0} and \Cref{fig:examples_1} we showcase a few randomly selected samples from the YCBV, HB, and LMO datasets using the noise discretizations $\rho=5$ and $\rho=15$. 
The top row of \Cref{fig:examples_0} shows a very common failure mode for YCBV. 
The clamps, of which there are multiple, only differ in size.
To to our template rendering scheme, where we crop closely around the foreground, this change in size is not reflected.
Hence, our model has no basis to compare and differentiate the two.

Another common failure mode can be seen in the bottom row of \Cref{fig:examples_1} on the LMO dataset. 
The object are generally quite small, most likely smaller than the average object size in our training data. 
This leads to some instances of the ensemble to estimate wrong instances which results in multiple instances per object in some cases.
These instances deteriorate the performance but since the objective of the coarse model is a high recall, we leave them in and let the refinement model compare which segmentation fits best to the model.

\begin{figure*}
  \centering
    \begin{subfigure}{0.2\linewidth}
    \includegraphics[width=\linewidth]{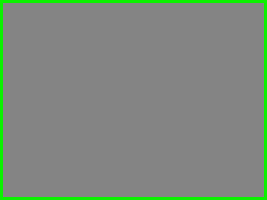}
    \caption{Full image}
  \end{subfigure}
  \hfill
  \begin{subfigure}{0.2\linewidth}
    \includegraphics[width=\linewidth]{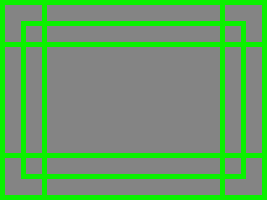}
    \caption{6 crops}
  \end{subfigure}
  \hfill
  \begin{subfigure}{0.2\linewidth}
    \includegraphics[width=\linewidth]{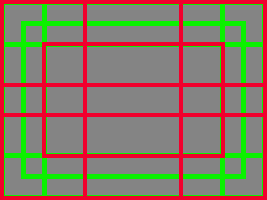}
    \caption{11 crops}
  \end{subfigure}
  \hfill
    \begin{subfigure}{0.2\linewidth}
    \includegraphics[width=\linewidth]{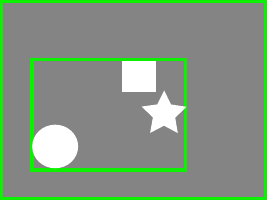}
    \caption{Full image guided}
  \end{subfigure}
  \caption{\textbf{Visualization of test-time spatial} augmentations. (a) uses the full image as input, (b) adds $5$ spatial crops to the full image, each smaller by a fixed number of pixels and shifted (c) adds another $5$ spatial augmentations by adding more crops smaller, and (d) illustrates the guided spatial augmentation. We use the previous model prediction to generate a tight bounding box covering all estimates that we run a diffusion for in a second step.}
  \label{fig:spatial}
\vspace{-1.5em}
\end{figure*}


\subsection{Training Datasets}
\label{}
To train our diffusion models, we generate two new datasets, one based on the 3D models of Google Scanned Objects~\cite{downs2022google} and a subset of objects from the Objaverse~\cite{deitke2023objaverse} dataset, picked from the lvis and staff-picked annotations.
We use Blenderproc~\cite{denninger2019blenderproc} for rendering and base the rendering pipeline closely to the scripts used to generate training datasets for the BOP challenge.
For each scene, we create a basic room environment and set up initial lighting from both a large ceiling plane and a point source. 
We randomly select a varying number of objects from our collection, load them into the virtual scene, and apply randomized material properties, such as roughness and specularity. 
To ensure realistic arrangements, these objects are enabled for physics simulation, allowing them to settle naturally on surfaces and against each other after an initial random placement. 
Lighting elements are also randomized in terms of color, strength, and position, and various textures are randomly applied to the room's surfaces to add visual diversity. 
Once the objects are settled, we generate multiple camera viewpoints for the scene, ensuring that the camera has an unobstructed and interesting view of the objects.
Finally, for each valid camera pose, it renders both color images and depth maps, along with annotations for the objects' positions and orientations, all of which are then saved in the BOP dataset format. 
After each scene is rendered and its data recorded, the objects are removed to prepare the environment for the next scene, allowing for the generation of a large number of unique synthetic scenarios.

\end{document}